\def\year{2021}\relax
\documentclass[letterpaper]{article} 
\usepackage{aaai21}  
\usepackage{times}  
\usepackage{helvet} 
\usepackage{courier}  
\usepackage[hyphens]{url}  
\usepackage{graphicx} 
\usepackage{natbib}  
\usepackage{caption} 
\usepackage{multirow}
\usepackage{ctable}
\usepackage{xcolor}
\usepackage{amsmath}
\usepackage{amssymb}
\newcommand{\dataName}{MedQA}
\renewcommand{\vec}[1]{\mathbf{#1}}
\title{What Disease does this Patient Have?\\
A Large-scale Open Domain Question Answering Dataset from Medical Exams}
\author {
    Di Jin,\textsuperscript{\rm 1}
    Eileen Pan,\textsuperscript{\rm 1}
    Nassim Oufattole\textsuperscript{\rm 1} \\
    Wei-Hung Weng,\textsuperscript{\rm 1}
    Hanyi Fang,\textsuperscript{\rm 2}
    Peter Szolovits\textsuperscript{\rm 1} \\
}
\affiliations {
    \textsuperscript{\rm 1} Computer Science and Artificial Intelligence, MIT, USA \\
    \textsuperscript{\rm 2} Tongji Medical College, HUST, PRC \\
    \{jindi15,eileenp,nassim,ckbjimmy,psz\}@mit.edu, fanghanyi@hust.edu.cn
}

\begin{document}

\maketitle

\begin{abstract}
Open domain question answering (OpenQA) tasks have been recently attracting more and more attention from the natural language processing (NLP) community. In this work, we present the first free-form multiple-choice OpenQA dataset for solving medical problems, \textsc{\dataName}, collected from the professional medical board exams. It covers three languages: English, simplified Chinese, and traditional Chinese, and contains 12,723, 34,251, and 14,123 questions for the three languages, respectively. We implement both rule-based and popular neural methods by sequentially combining a document retriever and a machine comprehension model. Through experiments, we find that even the current best method can only achieve 36.7\%, 42.0\%, and 70.1\% of test accuracy on the English, traditional Chinese, and simplified Chinese questions, respectively. We expect \textsc{\dataName} to present great challenges to existing OpenQA systems and hope that it can serve as a platform to promote much stronger OpenQA models from the NLP community in the future.\footnote{Data and baselines source code are available at: https://github.com/jind11/MedQA}
\end{abstract}

\section{Introduction}

Question answering (QA) is a fundamental task in Natural Language Processing (NLP), which requires models to answer a particular question. When given the context text associated with the question, language pre-training based models such as BERT \citep{devlin-etal-2019-bert}, RoBERTa \cite{Liu2019RoBERTaAR}, and ALBERT \cite{lan2019albert} have achieved nearly saturated performance on most of the popular datasets \cite{rajpurkar-etal-2016-squad,rajpurkar-etal-2018-know,lai-etal-2017-race,yang-etal-2018-hotpotqa,gao-etal-2020-machine}. However, real-world scenarios for QA are usually much more complex and one may not have a body of text already labeled as containing the
answer to the question. In this scenario, models are required to find and extract relevant information to questions from large-scale text sources such as a search engine \cite{Dunn2017SearchQAAN} and Wikipedia~\cite{chen-etal-2017-reading}. This type of task is generally called as open-domain question answering
(OpenQA), which has recently attracted lots of attention from the NLP community \citep{clark-gardner-2018-simple,Wang2019MultipassageBA,Asai2020LearningTR} but still remains far from being solved.

\begin{table*}[t]
\centering
\tiny
\begin{tabular}{>{\arraybackslash} m{1.2cm} |>{\arraybackslash} m{15.4cm}}
\toprule
\textbf{Question}                  & A 27-year-old male presents to urgent care complaining of pain with urination. He reports that the pain started 3 days ago. He has never experienced these symptoms before. He \textit{denies gross hematuria or pelvic pain}. He is sexually active with his girlfriend, and they consistently use condoms. When asked about recent travel, he admits to recently returning from a “boys' trip" in Cancun where he had \textit{unprotected sex} 1 night with a girl he met at a bar. The patient’s medical history includes type I diabetes that is controlled with an insulin pump. His mother has rheumatoid arthritis. The patient’s temperature is 99 F (37.2 C), blood pressure is 112/74 mmHg, and pulse is 81/min. On physical examination, there are no lesions of the penis or other body rashes. No costovertebral tenderness is appreciated. A urinalysis reveals no blood, glucose, ketones, or proteins but is \textit{positive for leukocyte esterase}. A urine microscopic evaluation shows a \textit{moderate number of white blood cells} but no casts or crystals. A urine culture is negative. Which of the following is the most likely cause for the patient's symptoms? \\ \midrule
\textbf{Options}                   & \textbf{A: Chlamydia trachomatis}, B: Systemic lupus erythematosus, C: Mycobacterium tuberculosis, D: Treponema pallidum \\ \midrule
\textbf{Evidence}                  & At least one-third of male patients with \textit{C. trachomatis} urethral infection have \textit{no evident signs or symptoms of urethritis}. ... Such patients generally have \textit{pyuria} ..., a \textit{positive leukocyte esterase test}, ... \\ \midrule \midrule
\textbf{Question}                  & A 57-year-old man presents to his primary care physician with a 2-month history of \textit{right upper and lower extremity weakness}. He noticed the weakness when he started falling far more frequently while running errands. Since then, he has had \textit{increasing difficulty} with walking and lifting objects. His past medical history is significant only for well-controlled hypertension, but he says that some members of his \textit{family have had musculoskeletal problems}. His right upper extremity shows \textit{forearm atrophy} and \textit{depressed reflexes} while his right lower extremity is \textit{hypertonic with a positive Babinski sign}. Which of the following is most likely associated with the cause of this patient’s symptoms? \\ \midrule
\textbf{Options}                   & A: HLA-B8 haplotype, B: HLA-DR2 haplotype, \textbf{C: Mutation in SOD1}, D: Mutation in SMN1, E: Viral infection \\ \midrule
\multirow{3}{*}{\textbf{Evidence}} & 1. The manifestations of ALS ...
\textit{insidiously} developing \textit{asymmetric weakness}, usually first evident distally in one of the limbs.\\
                          & 2. ... 
                          \textit{hyperactivity of the muscle-stretch reflexes (tendon jerks)} and, often, \textit{spastic resistance to passive movements} ...
                          \\
                          & 3. \textit{Familial ALS (FALS)}...
                          clinically indistinguishable from sporadic ALS... Genetic studies have identified mutations in multiple genes, including cytosolic enzyme \textit{SOD1}...
\\ \bottomrule
\end{tabular}
\caption{\label{tab:data-examples} Two examples of \textsc{\dataName}. The correct answer among options is marked in bold font. Key words in the question and evidence text to help answer the questions are highlighted in italic font. 
Evidence for both examples are from the textbook ``Harrison's Principles of Internal Medicine''.
}
\vspace*{-4mm}
\end{table*}

Most previous works for OpenQA focus on datasets in which answers are in the format of spans (several consecutive tokens) and can be found based on the information \textbf{explicitly} expressed in the provided text \cite{joshi-etal-2017-triviaqa,chen-etal-2017-reading,Dunn2017SearchQAAN,Dhingra2017QuasarDF}. As a more challenging task, free-form multiple-choice OpenQA datasets such as ARC \cite{Clark2018ThinkYH} and OpenBookQA \cite{mihaylov-etal-2018-suit} contain a significant percentage of questions focusing on the \textbf{implicitly} expressed facts, events, opinions, or emotions in the retrieved text. To answer these questions, models need to perform logical reasoning over the information presented in the retrieved text and in some cases even need to integrate some prior knowledge. Such datasets can motivate a general QA algorithm that can read any given question, find relevant evidence from a knowledge bank, and conduct logical reasoning to obtain the answer. Unfortunately, these OpenQA datasets consist of questions that require only elementary or middle school level knowledge (e.g., ``Which object would let the most heat travel through?''), so even excellent models trained on them may be unable to support more sophisticated real-world scenarios.

To this end, we introduce a new OpenQA dataset, \textsc{\dataName}, for solving medical problems, representing a demanding real-world scenario. Questions in this dataset are collected from medical board exams in US, Mainland China, and Taiwan, where human doctors are evaluated on their professional
knowledge and ability to make clinical decisions. Questions in these
exams are varied and generally require a deep understanding of related medical concepts learned from medical textbooks to answer. Table \ref{tab:data-examples} shows two typical examples. An OpenQA model must learn to find relevant information from the large collection of text materials we assembled from medical textbooks, reason over them, and make decisions about the right answer. Taking the first question in Table \ref{tab:data-examples} as instance, to obtain the correct answer ``Chlamydia trachomatis'', the OpenQA model first needs to retrieve the relevant evidence as shown in this table from a large collection of medical textbooks, read over the question body, pay attention to the key findings for this patient: no evident signs of \textit{urethritis}, finding of \textit{pyuria}, and positive \textit{leukocyte esterase test}, and infer the correct answer after extensive reasoning. 

To provide benchmarks for \textsc{\dataName}, we implement several state-of-the-art methods for the OpenQA task based on the standard system design, which consists of two components: a document retriever for finding relevant evidence text, and a document reader that performs machine comprehension over the retrieved evidence. Experimental results have shown that even the best method powered by the large pre-trained models~\cite{devlin-etal-2019-bert} can only achieve 36.7\%, 42.0\%, and 70.1\% of test accuracy on the questions collected from US, Mainland China, and Taiwan, respectively, indicating the great challenge of this dataset. Through both quantitative and qualitative analysis, we find that the performance of document retriever should be the bottleneck since its current form cannot conduct multi-hop reasoning over the retrieving process. Our hope is that \textsc{\dataName} can serve as a platform to encourage researchers to develop a general OpenQA model that can solve complex real-world questions via abundant logical reasoning in both retrieval and comprehension stages.

\section{Related Work}

\begin{table*}[]
\small
\resizebox{.98\textwidth}{!}{
\begin{tabular}{lcccccc|c}
\toprule
\textbf{Datasets}          & \textbf{LiveQA}          & \textbf{Medication QA}   & \textbf{BioASQ}     & \textbf{emrQA}                      & \textbf{MEDIQA}          & \textbf{MedQuAD}         & \textbf{\dataName}                \\ \midrule
\textbf{Question Source}   & consumer health & consumer health & biomedical         & EMR & consumer health & consumer health & medical\&clinical        \\
\textbf{Answer Format}     & retrieval       & retrieval       & span based\&binary & ranking                  & ranking         & retrieval       & multiple choice \\
\textbf{Annotation Method} & manual          & manual          & manual             & automatic                  & manual          & automatic       & manual          \\
\textbf{Dataset Size}      & 660             & 674             & 3,743              & 455,837                    & 383             & 47,457          & 61,097         \\ 
\textbf{OpenQA?} & No & No& No& No& No& No& Yes \\
\bottomrule
\end{tabular}}
\caption{Comparison of our dataset with existing medical QA datasets. 
In terms of the answer format, ``retrieval'' means that the answer is a snippet of retrieval result by searching relevant websites; ``ranking'' means that the answer is a list of candidates and the task is to rank them with higher ranking ones being better answers. Automatic annotation is obtained via an algorithm.
}
\label{tab:datasets-comparison}
\vspace*{-4mm}
\end{table*}

Traditionally, QA tasks have been designed to be text-dependent~\cite{rajpurkar-etal-2016-squad,lai-etal-2017-race,reddy2019coqa}, where a model is required to comprehend a given text to answer questions. Those given texts relevant to the questions are specially curated by people, which is infeasible for real-world applications where annotations of relevant context are expensive to obtain. This gave rise to the birth of open-domain QA (OpenQA) task, where models must both find and comprehend the context for answering the questions. As a preliminary trial, \citet{chen-etal-2017-reading} proposes the DrQA model, which uses a text retriever to obtain relevant documents from Wikipedia, and further applies a trained reading comprehension model to extract the answer from the retrieved documents. Afterwards, researchers have introduced more sophisticated models, which either aggregate all informative evidence \cite{clark-gardner-2018-simple,Wang2019MultipassageBA} or filter out those irrelevant retrieved texts \cite{Wang2018R3RR,Das2019MultistepRI} to better predict the answers. Benefiting from the power of neural networks, these models have achieved remarkable progress in OpenQA.

Much of the previous OpenQA work focuses on the datasets whose answers are spans from the retrieved documents~\cite{joshi-etal-2017-triviaqa,chen-etal-2017-reading,Dunn2017SearchQAAN}. In general, most questions concern the facts that are explicitly expressed in the text, 
offering an advantage to systems that rely mostly on surface word matching~\cite{Sun2019InvestigatingPK}. 

To promote more advanced reading skills, another research line has studied OpenQA tasks in a free-form multiple-choice form~\cite{Clark2018ThinkYH,mihaylov-etal-2018-suit}.
These benchmarks allow a relatively more comprehensive evaluation of different higher-level reading skills such as logical reasoning and prior knowledge integration. In particular, real-world exams such as SAT and Gaokao are ideal sources for constructing this kind of OpenQA datasets~\cite{Clark2018ThinkYH,mihaylov-etal-2018-suit,zhang2018medical}.
Our proposed dataset follows this line but differs in three aspects:

\begin{itemize}
\item The source of our dataset is designed to examine the doctors' professional capability and thus contains a significant number of questions that require multi-hop logical reasoning, which helps push the development of reading comprehension models along this direction~\cite{Yu2020ReClorAR}. 
\item Our dataset is the first publicly available large-scale multiple-choice OpenQA dataset for the medical problems, where extensive prior domain-specific knowledge is anticipated for the model. It can thus contribute to the emerging field where a general language model will need to be combined with world knowledge. 
\item Our dataset is cross-lingual, covering English and simplified/traditional Chinese, which contributes to the emerging field of cross-lingual natural language understanding.
\end{itemize}

There are several related medical QA datasets, which are summarized in Table \ref{tab:datasets-comparison}. Among them, LiveQA~\citep{abacha2017overview}, Medication QA~\citep{abacha2019bridging}, MEDIQA~\citep{ben-abacha-etal-2019-overview}, and MedQuAD~\citep{abacha2019question} contain consumer health related questions and the answers are obtained by searching healthcare websites such as MedlinePlus
via the ChiQA system~\citep{demner2020consumer} and selecting a snippet of searching results out, where the answer retrieval is performed by keywords matching and complex reasoning is seldom involved. BioASQ~\citep{nentidis2020overview} is similar to the SQuAQ dataset~\citep{rajpurkar-etal-2016-squad}, where a span of text in the given context is used as the answer and thus no external knowledge source is needed. emrQA~\citep{pampari-etal-2018-emrqa} aims to rank a list of Electronic Medical Record (EMR) text lines to find the best line as the answer and the ground truth answers are obtained via an algorithm rather than human annotation. Overall, none of these related datasets have been formulated as an OpenQA problem. Last but not the least, \citet{zhang2018medical} and \citet{ha-yaneva-2019-automatic} have previously worked on the Chinese and English versions of our proposed dataset, respectively. However, the former one did not release any data while the latter one only released 454 questions for public use, while we publicize a large-scale dataset to promote more powerful deep models. 

\section{Data}

\subsection{Task Formulation}

The task is defined by its three components:

\paragraph{Question:} 
question in text, either in one sentence asking for a certain piece of knowledge, or in a long paragraph starting with a description of the patient condition.
\paragraph{Answer candidates:} multiple answer options are given for each question, of which only one should be chosen as the most appropriate.
\paragraph{Document collection:} a collection of text material extracted from a variety of sources and organized into paragraphs, which contains the knowledge and information to help find the answers.

This task is to determine the best answer to the question among the candidates, relying on the documents.

\subsection{Data Collection}

\paragraph{Questions and Answers} 
We collected the questions and their associated answer candidates from the National Medical Board Examination in the USA
, Mainland China
, and Taiwan
. For convenience, we denote the datasets from these three sources as USMLE, MCMLE, and TWMLE, respectively. These tests assess a physician's ability to apply knowledge, concepts, and principles, and the ability to demonstrate fundamental patient-centered skills. 
We include problems from both real exams and mock tests; all are freely accessible online for public usage. Details about the sources that we collected data from are described in Table A.1 of the Supplementary Material. We also provided the scripts of data collection and processing in the Supplementary Material.

We remove duplicate problems and randomly split the data based on questions, with 80\% training, 10\% development, and 10\% test. The overall statistics of the dataset are summarized in Table \ref{tab:data-stats}. 
To comply with fair use of law\footnote{https://www.copyright.gov/fair-use/more-info.html}, we shuffle the order of answer options and randomly delete one of the wrong options for each question for USMLE and MCMLE datasets, which results in four options with one right option and three wrong options.
Percentages of each option as the correct answer for the development set are summarized in Table A.2 of the Supplementary Material.

\begin{table}[!htpb]
\centering
\small
\resizebox{.48\textwidth}{!}{
\begin{tabular}{lrrr}
\toprule
\textbf{Metric}            & \textbf{USMLE} & \textbf{MCMLE} & \textbf{TWMLE} \\ \midrule
\# of options per question & 4            & 4                       & 4               \\ 
Avg./Max. option len.    & 3.5 / 45     & 7.3 / 100               & 20.6 / 210      \\ 
Avg./Max. question len.  & 116.6 / 530  & 45.7 / 333              & 61.0 / 1950     \\ 
vocab/character size            & 63317        & 3263                    & 3588            \\ \midrule
\multicolumn{4}{l}{\textbf{\# of questions}}                                                   \\ 
Train                      & 10178        & 27400                   & 11298           \\ 
Development                & 1272         & 3425                    & 1412            \\ 
Test                       & 1273         & 3426                    & 1413            \\ 
All                        & 12723        & 34251                   & 14123            \\ \toprule
\end{tabular}}
\caption{\label{tab:data-stats} Overall statistics of \textsc{\dataName}. 
Question/option length and vocabulary/character size are calculated in tokens for English and in characters for Chinese. Vocabulary/character size is measured on the combination of questions and options. ``Avg./Max. option len.'' represents ``Average/Maximum option length''.}
\end{table}

\paragraph{Document Collection}
Extensive medical knowledge is needed to answer every question in our \textsc{\dataName} data. For people to obtain answers for these questions, they need to obtain necessary knowledge from a volume of medical textbooks during years of training. Similarly, for a machine learning model to be successful in this task, we need to grant it access to the same collection of text materials as human have. Therefore, for USMLE, we prepared text materials from a total of 18 English medical textbooks that have been widely used by medical students and USMLE takers, whereas for MCMLE, we collected 33 simplified Chinese medical textbooks designated as the official textbooks for preparing the medical licensing exam in Mainland China. For TWMLE, since medical students in Taiwan use the same textbooks as those in USA for exam preparation, USMLE and TWMLE would use the same document collection for solving questions. We will release the textbooks we collected upon the license agreement of research use only.

All textbooks we collected are originally in PDF format and we converted them into digital text via OCR.
We performed some clean-up pre-processing over the converted text such as misspelling correction, and then divided all text into paragraphs. Table \ref{tab:document-stats} summarizes the statistics of the document collection.

\begin{table}[!htpb]
\small
\centering
\resizebox{.48\textwidth}{!}{
\begin{tabular}{lrr}
\toprule
\textbf{Metric}                 & \textbf{USMLE/TWMLE}       & \textbf{MCMLE}        \\ \midrule
\# of books            & 18          & 33           \\
\# of paragraphs       & 231,581     & 116,216       \\
\# of tokens           & 12,727,711  & 14,730,364     \\
Vocabulary size             & 245,851     & 4,695         \\
Avg./Max. paragraph length & 55.0 / 1,234 & 126.7 / 9,082 \\\bottomrule
\end{tabular}}
\caption{\label{tab:document-stats} Overall statistics of the document collection. USMLE and TWMLE share the same document collection. Token number and vocabulary size are counted in tokens for English and in characters for Chinese.}
\end{table}

To evaluate whether these collected text materials can cover enough knowledge to answer those questions, we randomly extracted 100 questions from the development set of all three datasets and let two medical experts with the MD degree annotate how many of them can be answered by the evidence from our prepared text materials. Table \ref{tab:document-coverage} summarizes the results of this evaluation. From this table, we see that our collected text materials can provide enough information for answering most of the questions in our data.

\begin{table}[]
\centering
\small
\begin{tabular}{lrrr}
\toprule
              & \textbf{USMLE} & \textbf{MCMLE} & \textbf{TWMLE} \\\midrule
Coverage (\%) & 88.0             & 100.0            & 87.0 \\
\bottomrule
\end{tabular}
\caption{\label{tab:document-coverage} Percentage of questions that human experts can find enough evidence from our collected text material to obtain correct answers for by randomly annotating 100 samples from the development set.}
\end{table}

\subsection{Data Analysis}

Based on a preliminary analysis of our proposed data, we find that it poses unique challenges for language understanding compared with existing OpenQA datasets, elaborated below:

    \paragraph{Professional Knowledge:} For most existing QA datasets, the question answering process relies largely on the basic understanding of language and general world knowledge reasoning. Some works have revealed that the large-scale pretrained language models bear a certain level of common-sense and symbolic reasoning capability besides their linguistic knowledge \cite{Talmor2019oLMpicsO,Petroni2019LanguageMA}, which may contribute significantly to the remarkable performance of current QA models. 
    However, answering of every question in our dataset needs abundant professional domain-specific knowledge, particularly medical knowledge, which forces the model to have a deep understanding of the extracted context. 
    \paragraph{Diversity of Questions:} The field of clinical medicine is diverse enough for questions to be asked about a wide range of topics. In general, there are two categories of questions: 1). The question is asking for a single piece of knowledge, for instance via the question "Which of the following symptoms belongs to schizophrenia?" 2). The question first describes a patient’s condition and then asks for \textit{the most probable diagnosis / the most appropriate treatment / the examination needed / the mechanism of certain conditions / the possible outcome for a certain treatment}, etc. Table \ref{tab:data-examples} shows two typical examples for type 2. Table \ref{tab:question-types} summarizes the percentages of these two types of questions for each dataset by annotating randomly selected 100 questions from the development set. Typically, type 1 questions need one-step reasoning while type 2 questions require multi-hop reasoning and are thus much more complicated than type 1 ones, imposing challenges not only to the reading comprehension model but also to the relevant text retrieval module. For example, in order to solve the first question in Table \ref{tab:data-examples}, the model needs to first extract, understand, and interpret the symptoms of this patient among a long paragraph of description,  then match these symptoms to millions of medical knowledge text snippets and find out the most relevant one, and finally understand the evidence sentence for answer selection.
    \paragraph{Complex Reasoning over Multiple Evidence:} Many questions in our data 
    involve complex multi-hop reasoning over several evidence snippets.  For instance, the second example is a typical question that requires multiple steps of reasoning over three evidences, where from the symptoms and signs we can know that this patient is highly likely an ALS case by looking at the evidence 1 and 2. Afterwards, from evidence 3, we know that the SOD1 is the possible genetic mutation for familial ALS, which is the correct answer. 
    \paragraph{Noisy Evidence Retrieval:} Retrieving relevant information from large-scale text is much more challenging than reading a short piece of text. Passages from textbooks often do not directly give answers to questions and many passages retrieved by the most widely adopted term-matching based information retrieval (IR) systems turn out to be noisy distractors and not relevant, especially for type 2 questions. For those questions involving multi-hop reasoning, models must identify all relevant information scattered in different passages, and missing any single piece of evidence would lead to failures. 

\begin{table}[!htpb]
\centering
\small
\begin{tabular}{lrrr}
\toprule
\textbf{Question Types} & \textbf{USMLE} & \textbf{MCMLE} & \textbf{TWMLE} \\\midrule
Type 1                 & 2.0            & 75.0           &     69.0           \\
Type 2                 & 98.0          & 25.0           &       31.0        \\\bottomrule
\end{tabular}
\caption{\label{tab:question-types} Percentages of type 1 and type 2 questions by annotating 100 randomly samples data from the development set. Type 1 questions ask about a single knowledge point, whereas type 2 questions simulate the realistic clinical settings by studying a patient case.}
\end{table}

\section{Approaches}

We implement both classical rule-based methods and recent state-of-the-art neural network based models. 

\subsection{Rule-based Methods}

We first propose two rule-based methods that do not involve a training process.

\subsubsection{Pointwise Mutual Information (PMI)}
This method is based on the PMI score function~\cite{clark2016combining}, which measures the strength of association between two n-grams $x$ and $y$ and is defined as:

\begin{equation*}
    PMI(x, y)=log\frac{p(x)p(y)}{p(x,y)},
\end{equation*}
where $p(x, y)$ is the joint probability that $x$ and $y$ occur together in our document collection $C$, within a certain window of text (we use a 10 word window); $p(x)$ is the probability that $x$ occurs in $C$. In practice, we use the frequency to represent the probability.
The larger this PMI score, the stronger the association between $x$ and $y$.

This method extracts unigrams, bigrams, trigrams, and skip-bigrams from the question $q$ and each answer option $a_i$ and calculates the average PMI score over all pairs of question n-grams and answer option n-grams. The answer option with the highest average PMI score will be picked as the prediction.

\subsubsection{Information Retrieval (IR)}
\label{sec:IR}

As the preliminary trial, we adopt a standard off-the-shelf text retrieval system built upon Apache Lucene, Elasticsearch,
using inverted index lookup followed by BM25 ranking. Specifically, for each question $q$ and each answer option $a_i$, we send $q + a_i$ as a query to the search engine, and return the search engine’s score for the top-N retrieved sentences. This is repeated for all options to score them all, and the option with the highest score is selected. We denote this version as \textsc{IR-ES}.

To seek for better IR performance, we enhance the above using a BM25 re-weighting mechanism inspired by~\citet{chen-etal-2017-reading}. The updated scoring function is detailed in Section 2 of the Supplementary Material due to page limits. Our best performing system uses unigram counts.
For English questions, we perform Snowball stemming to both documents and questions; and we use the MetaMap\footnote{https://metamap.nlm.nih.gov/} tool to identify and remove the non-medically-related words out of the questions since they are not useful for medical evidence retrieval. Details of this step can be found in the Section C of the Supplementary Material. We denote this version as \textsc{IR-Custom}.

\subsection{Neural Models}

Following the DrQA system by \citet{chen-etal-2017-reading}, this line of models consists of two components: (1) the Document Retriever module for finding relevant passages and (2) a machine comprehension model, Document Reader, for obtaining the answer by reading the small collection of passages. 

\subsubsection{Document Retriever}

We use the best IR system developed in Section \ref{sec:IR} and obtain the top-N ranked passages from the large-scale document collection $C$, concatenating them into a long sequence $c$. Then for each question and option pair $qa_i=q+a_i$, $qa_i$ and $c$ are then passed to the Document Reader for reasoning and decision-making.

\subsubsection{Document Reader}

We implement the following widely used document reader models:

\paragraph{\textsc{Max-out}:}
Following~\citet{mihaylov-etal-2018-suit}, we first use the same bi-directional gated recurrent unit (BiGRU) model to encode both the context $c$ and the question and option pair $qa_i$, and then perform max-pooling to obtain the final representation vectors $\vec{h_c}\in \mathbb{R}^h$ and $\vec{h_{qa_i}}$. We then use the following equation to calculate the probability score of how likely this option $a_i$ is to answer this question $q$ given the retrieved context $c$:

\begin{align*}
\small
    &\Vec{h}=[\vec{h_c};\vec{h_{qa_i}};\vec{h_c}\cdot\vec{h_{qa_i}};|\vec{h_c}-\vec{h_{qa_i}}|],\\
    &p(q, a_i|c)=W_1(tanh(W_2\vec{h}))\in \mathbb{R}^1,
\end{align*}
where $[\cdot;\cdot]$ represents the concatenation operation; $W_1\in \mathbb{R}^{1\times h}$ and $W_2\in \mathbb{R}^{h\times 4h}$ are weight matrices to be learned. We compute such a probability score for each option and select the option with the highest score.



\paragraph{Fine-Tuning Pre-Trained Language Models:}
We also apply the framework of fine-tuning a pretrained language model such as BERT~\cite{devlin-etal-2019-bert} on our data following~\citet{Jin2019MMMMM,yan-etal-2020-multi}. Specifically, we construct the input sequence by concatenating [CLS], tokens in $c$, [SEP], tokens in $qa_i$, [SEP], where [CLS] and [SEP] are the classifier token and sentence separator in a pre-trained language model, respectively. We denote the hidden state output by the first token as $\Vec{h}\in \mathbb{R}^h$ and obtain the unnormalized
log probability $p(q, a_i|c)=W\Vec{h}\in \mathbb{R}^1$, where $W\in \mathbb{R}^{1\times h}$ is the weight matrix. We obtain the final prediction by applying a softmax layer over the unnormalized log probabilities of all options associated with $q$ and picking the option with the highest probability.

\section{Experiments}

\subsection{Experimental Settings}
\label{sec:exp-settings}

\paragraph{Dealing with TWMLE Data:}
Since TWMLE uses the same document collection as USMLE for solving problems, we translate the questions in TWMLE from traditional Chinese to English via Google Translation
and then use the same models as USMLE.

\paragraph{\textsc{Max-out}:}
We use spaCy
as the English tokenizer and HanLP
as the Chinese tokenizer. We use the 200-dimensional word2vec word embeddings induced from PubMed and PMC texts~\cite{moen2013distributional} for English text,
and use the 300-dimensional Chinese fastText word embeddings
~\cite{bojanowski-etal-2017-enriching}.
The maximum sequence length for a passage is limited to 450 tokens while that for the question and answer pair is limited to 150 tokens.

\paragraph{Fine-Tuning Pre-Trained Language Models:}
We use the following pre-trained language models for Chinese: Chinese BERT-Base (denoted as
\textsc{BERT-Base-Zh}) released by Google~\cite{devlin-etal-2019-bert},
Chinese BERT-Base with whole word masking during pre-training over larger corpora (denoted as \textsc{BERT-Base-wwm-ext})~\cite{chinese-bert-wwm},
multilingual BERT-Base (uncased, denoted as \textsc{MBERT-Base})~\cite{devlin-etal-2019-bert},
Chinese RoBERTa-Large with whole word masking over the same corpora as BERT-Base-wwm-ext (denoted as \textsc{RoBERTa-Large-wwm-ext})~\cite{chinese-bert-wwm}. 

For English, we consider the following pre-trained models: English BERT-Base (denoted as \textsc{BERT-Base-En})~\cite{devlin-etal-2019-bert}, English BioBERT-Base and BioBERT-Large that fine-tune the English BERT models further over the bio-medical literature from PubMed (denoted as \textsc{BioBERT-Base/Large})~\cite{10.1093/bioinformatics/btz682},
English Clinical BERT-Base that fine-tunes the BERT-Base model further over the clinical notes extracted from MIMIC-III (denoted as \textsc{clinicalBERT})~\cite{alsentzer-etal-2019-publicly},
bio-medical RoBERTa-Base by adapting RoBERTa-Base to bio-medical scientific papers from the Semantic Scholar corpus (denoted as \textsc{BioRoBERTa-Base})~\cite{domains},
English RoBERTa-Large (\textsc{RoBERTa-Large})~\cite{Liu2019RoBERTaAR}.
We did not try further fine-tuning BERT models on our collected textbooks since they only contain 12-15 M tokens, which are far less than the typical corpus size for model pre-training (containing over billions of tokens).

We set the learning rate and effective batch size (product of batch size and gradient accumulation steps) to $2\times10^{-5}$ and 18 for base models, and $1\times10^{-5}$ and 6 for large models. We truncate the longest sequence to 512 tokens after sentence-piece tokenization (we only truncate context). We fine-tune English/Chinese models for 8/16 epochs, set the warm-up steps to 1000, and keep the default values for the other hyper-parameters~\cite{devlin-etal-2019-bert}.

\begin{table}[!htpb]
\centering
\small
\begin{tabular}{lrr}
\toprule
\textbf{Methods}      & \textbf{Dev} & \textbf{Test} \\ \midrule
\textsc{Chance}  &     25.0 & 25.0     \\ \midrule
\textsc{PMI}          &  36.6   & 36.9     \\
\textsc{IR-ES}        &  38.3   &   37.2   \\
\textsc{IR-Custom}    &  39.1   &  37.8    \\ \midrule
\textsc{Max-out}      &  51.8   &  50.9    \\
\midrule
\textsc{BERT-Base-Zh} & 66.5    & 65.8 \\
\textsc{BERT-Base-wwm-ext}   &  64.4   & 64.0      \\
\textsc{MBERT-Base}             &  62.1   & 62.3     \\ \midrule
\textsc{RoBERTa-Large-wwm-ext}             &  \textbf{69.3}   & \textbf{70.1}    \\ \bottomrule
\end{tabular}
\caption{Performance of baselines in accuracy (\%) on the MCMLE dataset.}
\label{tab:MCMLE-results}
\end{table}

\subsection{Baseline Results}

Tables \ref{tab:MCMLE-results} and \ref{tab:USMLE-results} summarize the performance of all baselines on the three datasets. By looking at these two tables, we observe:

\begin{itemize}
    \item Our customized IR system outperforms the off-the-shelf version and the improvement is larger on the English questions.
    \item For the MCMLE dataset, pretrained models significantly outperform non-pretrained models and all neural models outperform the non-neural baselines. However, for the USMLE and TWMLE datasets, the non-pretrained neural models (\textsc{Max-out}) cannot even surpass the IR baseline, and most surprisingly, most of the pretrained models for the USMLE dataset cannot even beat the IR baseline. 
\end{itemize}

Overall, even the strongest pretrained model (\textsc{BioBERT-Large}, \textsc{RoBERTa-Large}) cannot harvest good scores on any of the three datasets, validating the great challenge of our proposed data. Notably, we did not include human performance for comparison due to the high variance of scores for human examinees (best medical students can earn almost full marks while worse ones cannot pass the exams). 

\begin{table}[!htpb]
\centering
\small
\begin{tabular}{lcccc}
\toprule
\multirow{2}{*}{\textbf{Datasets}}      & \multicolumn{2}{c}{\textbf{USMLE}} & \multicolumn{2}{c}{\textbf{TWMLE}} \\
& \textbf{Dev}         & \textbf{Test}        & \textbf{Dev}         & \textbf{Test}        \\\midrule
\textsc{Chance}  &     25.0 & 25.0 &     25.0 & 25.0     \\ \midrule
\textsc{PMI}          &  29.8   & 31.1 & 30.8 & 31.1     \\
\textsc{IR-ES}        &  34.0   & 35.5   & 24.9 & 26.8   \\
\textsc{IR-Custom}    &  \textbf{38.3}   & 36.1 & 35.1 & 34.8     \\ \midrule
\textsc{Max-out}      & 28.9  & 28.6 &  29.4  & 27.8 \\
\midrule
\textsc{BERT-Base-En} &  33.9 & 34.3 &  34.3 &33.3 \\
\textsc{clinicalBERT-Base} & 33.7  & 32.4 & 33.4 & 32.1 \\
\textsc{BioRoBERTa-Base} & 35.1 & 36.1 & 38.7 &36.9  \\
\textsc{BioBERT-Base}   & 34.3  & 34.1 & 41.6 & 41.1     \\ \midrule
\textsc{RoBERTa-Large} & 35.2 & 35.0 &  39.6 & 39.3 \\ 
\textsc{BioBERT-Large}   & 36.1  & \textbf{36.7} &   \textbf{42.2} & \textbf{42.0}     \\
\bottomrule
\end{tabular}
\caption{Performance of baselines in accuracy (\%) on the USMLE and TWMLE datasets.}
\label{tab:USMLE-results}
\vspace*{-4mm}
\end{table}

\subsection{Error Analysis}

\subsubsection{Quantitative Analysis}

Since our approaches to the proposed data involve two stages, i.e., document retrieval and reading comprehension, both stages could be the potential error sources. We first check the performance of the IR based document retrieval stage by letting two medical doctors annotate whether the top 25 retrieved paragraphs by \textsc{IR-Cust} contain enough evidence for answering the questions over 100 randomly selected samples from the development set.
We have three annotation levels: full evidence (evidence is totally complete to derive the answer), partial evidence (evidence is useful but not complete), and no evidence (no evidence is found at all). Table \ref{tab:evidence} summarizes the results of the annotation. We also summarize the percentages of samples for which we find full, partial, or no evidence individually for type 1 and type 2 questions on the MCMLE and TWMLE datasets. From this table, we see that only the MCMLE dataset has a good retrieval recall while for the majority of samples in USMLE, we cannot find any evidence in the retrieved text. Moreover, among those samples that can find full or partial evidence, we calculate the percentage of samples that find the evidence in the top $N$ (1, 5, 10, 15) retrieved paragraphs, as shown in Table \ref{tab:evidence-topN}. Overall, we have two findings: 1). The MCMLE and TWMLE datasets have much higher percentage of type 1 questions than USMLE, which is positively correlated with their much better IR retrieval performance; 2). Such poor evidence retrieval performance for the USMLE dataset should be the main cause of the extremely low baseline models' performance as shown in Table \ref{tab:USMLE-results} (neural models are even beaten by the IR baseline).

\begin{table}[!htpb]
\centering
\small
\resizebox{1.0\columnwidth}{!}{
\begin{tabular}{lrrr}
\toprule
\textbf{Types}            & \textbf{USMLE} & \textbf{MCMLE} & \textbf{TWMLE} \\ \midrule
Full Evidence    &  24.0     &    75.0 (82.7/52.0)   & 60.0 (63.4/52.6)      \\
Partial Evidence &  8.0     &    21.0 (14.7/40.0)  &  16.7 (19.5/10.5)    \\
No Evidence      &  68.0     &     4.0 (2.6/8.0) &  23.3 (17.1/36.9) \\ \bottomrule    
\end{tabular}}
\caption{Percentage of samples that can find full evidence, partial evidence, or no evidence in the top 25 retrieved paragraphs. For numbers within parenthesis, the left number is for type 1 while the right one is for type 2 questions. Since almost all questions are type 2 for USMLE, we do not have percentage numbers specifically for type 1 and 2 questions.}
\label{tab:evidence}
\end{table}

\begin{table}[!htpb]
\centering
\small
\begin{tabular}{lrrrr}
\toprule
\textbf{Datasets} & \textbf{Top 1} & \textbf{Top 5} & \textbf{Top 10} & \textbf{Top 15} \\ \midrule
USMLE    &  0.0     & 31.2  & 56.2   & 81.2   \\
MCMLE    &  66.7     & 92.7  & 96.9   & 100.0   \\
TWMLE    &  0.0     & 71.7  &  91.3  & 93.5   \\ \bottomrule
\end{tabular}
\caption{Percentage of samples that can find evidence in the top $N$ (1, 5, 10, 15) retrieved paragraphs among samples that can find full or partial evidence.}
\label{tab:evidence-topN}
\vspace*{-4mm}
\end{table}

\subsubsection{Qualitative Analysis}

Seeing such poor retrieval performance on the USMLE dataset, we wonder what could be the reasons. After taking a close look into the successful and failed samples in the retrieval stage, we can summarize one success pattern as well as two failure patterns for the majority of cases, described in the sections below. Remember that almost all questions in USMLE are about case studies (type 2 questions). 
And the IR system always returns us snippets with each of which matching to only a small portion of the question text. 

\paragraph{Success Patterns:}
The question asks about the most probable diagnosis, which involves only one step of reasoning (inference from symptoms / signs / findings to diagnosis of diseases), and it is easy to constrain the possible diagnosis candidates to one or two based on some special condition terms. In this case, it is easy for the IR system to obtain useful evidence snippets by matching those key condition terms. For example, given a case of a stressful young female patient suffering from \textit{recurrent headaches} that \textit{alternately affect the right or left brain} and are \textit{exacerbated by loud sounds or bright light}, along with \textit{nausea}, the IR system can identify these typical, condition-specific terms (highlighted by italic font) and retrieve the suitable evidence about the disease of \textit{migraine headache}.

\paragraph{Failure Patterns:}
1). The question still asks about the most probable diagnosis, however, each of the patient's symptoms is very common and could correspond to many possible diagnoses. In this case, the IR system may return miscellaneous diseases' descriptions, each of which can match part of the patient's symptoms, signs or other findings. However, none of these retrieved texts are relevant to the correct diagnosis. We provide one example in Section D of the Supplementary Material for illustration. 2). The question asks about the \textit{the most appropriate treatment / the examination needed / the mechanism of certain condition}, etc., which all involve two-steps of reasoning. That is, we need to first derive the diagnosis based on the patient's condition and then answer this question based on the inferred diagnosis. In this case, it is highly likely that the IR system only return evidence that enables the first step of reasoning (making the diagnosis) but not the second one. We also provide an example in Section D of the Supplementary Material. 

Last but not the least, we did a qualitative analysis of the translation quality from traditional Chinese to English for TWMLE questions during experiments and found that most of the sentences are good enough while a few of them look not that fluent. However, since all medical terms in TWMLE questions are originally in English, such minor in-fluency would not affect evidence retrieval and question answering.

\section{Conclusion}

We present the first open-domain multiple-choice question answering dataset for solving medical problems, \textsc{\dataName}, collected from the real-world professional examinations, requiring extensive and advanced domain knowledge to answer questions.
This dataset covers three languages: English, simplified Chinese, and traditional Chinese. 
Together with the question data, we also collect and release a large-scale corpus from medical textbooks from which the reading comprehension models can obtain necessary knowledge for answering the questions. 
We implement several state-of-the-art methods as baselines to this dataset by cascading two components: document retrieval and reading comprehension. And experimental results demonstrate that even current best approach cannot achieve good performance on these data. We anticipate more research efforts from the community can be devoted to this dataset so that future OpenQA models can be strong enough to solve such real-world complex problems.

\bibliography{bibfile}

\appendix
\renewcommand*\thetable{\Alph{section}.\arabic{table}}

\section{Data Collection}

For USMLE and MCMLE datasets, we scraped several websites that provide the question banks, and for TWMLE dataset, we downloaded the examination materials in PDF format directly from the official National Examination website in Taiwan and then converted them into digital format via Optical Character Recognition (OCR). Table \ref{tab:sources-stats} lists the websites that we obtained data from. 

\begin{table}[!htpb]
\small
\centering
\begin{tabular}{ll}
\toprule
\textbf{Datasets} & \textbf{Websites}                                                                                                                                       \\ \midrule
USMLE    & \begin{tabular}[c]{@{}l@{}}https://step1.medbullets.com\\ https://www.amboss.com/us/usmle\\ https://www.lecturio.com/usmle-step-1\end{tabular} \\\midrule
MCMLE    & http://www.offcn.com/yixue/linchuang\\\midrule
TWMLE    & \begin{tabular}[c]{@{}l@{}}https://wwwq.moex.gov.tw/exam/\\wFrmExamQandASearch.aspx\end{tabular} \\
\bottomrule
\end{tabular}
\caption{\label{tab:sources-stats} Websites that we collected our data from.}
\end{table}

\begin{table}[!hptb]
\small
\centering
\begin{tabular}{lrrr}
\toprule
\textbf{Options} & \textbf{USMLE}  & \textbf{MCMLE} & \textbf{TWMLE} \\ \midrule
A       & 22.5 & 25.9           & 20.3   \\ 
B       & 27.1 & 24.8           & 25.8   \\ 
C       & 26.1 & 27.7           & 29.0   \\ 
D       & 24.3 & 21.5           & 24.9   \\ \bottomrule
\end{tabular}
\caption{\label{tab:options-stats} Percentages of each option as the correct answer for the development sets.}
\end{table}

\section{Information Retrieval (IR)}

In this section we describe how the BM25 re-weighted BM25 scoring function works using the following equation:

\begin{small}
\begin{equation*}
    s(Q, D)=\sum_{i=1}^{n}\frac{BM_{25}(q_i, D)\cdot IDF(q_i)\cdot f(q_i, Q)\cdot (k_Q+1)}{f(q_i, Q)+k_Q\cdot(1-b_Q+b_Q\cdot\frac{queryLen}{avgQueryLen})},
\end{equation*}
\end{small}

\begin{small}
\begin{equation*}
    BM_{25}(q_i, D)=\frac{IDF(q_i)\cdot f(q_i, D)\cdot (k_D+1)}{f(q_i, Q)+k_D\cdot(1-b_D+b_D\cdot\frac{docLen}{avgDocLen})},
\end{equation*}
\end{small}
where $Q=\{q_1,q_2,...,q_n\}$ is the query consisting of $n$ query terms (each query term in our method is actually a n-gram); $IDF(q_i)$ the inverse document frequency of the query term $q_i$; $f(q_i,Q)$ and $f(q_i,D)$ represents the frequency of query term $q_i$ in the query $Q$ and the document $D$, respectively; $k_D$, $k_Q$, $b_D$, and $b_Q$ are hyper-parameters and their values in our experiments are summarized in the Table \ref{tab:k_b}; $queryLen$ and $docLen$ are the length in tokens of the current query $Q$ and document $D$, respectively; $avgQueryLen$ and $avgDocLen$ are the average length of all queries and documents.

\begin{table}[!htbp]
\centering
\begin{tabular}{lrrrr}
\toprule
\textbf{Datasets} & \textbf{$k_Q$} & \textbf{$b_Q$} & \textbf{$k_D$} & \textbf{$b_D$} \\ \midrule
USMLE             & 0.40          & 0.70          & 0.90          & 0.35          \\
MCMLE             & 1.10          & 0.20          & 0.90          & 0.35          \\
TWMLE             & 0.90          & 1.00          & 0.90          & 0.35 
\\ \bottomrule
\end{tabular}
\caption{Hyper-parameters for the IR system tuned on the development set.}
\label{tab:k_b}
\end{table}

\section{MetaMap for IR}

Since type 2 questions are long and contain many words that are not useful for retrieving the relevant context, we adopt the MetaMap tool to filter out those unwanted words. MetaMap was developed by the National Library of Medicine (NLM) to map biomedical text to concepts in the Unified Medical Language System (UMLS). It uses a hybrid approach combining a natural language processing (NLP), knowledge-intensive approach and computational linguistic techniques. We can use it to extract and standardize medical concepts from any biomedical/clinical text. Table \ref{tab:metamap-example} shows one example by sending one question to MetaMap API and parsing the returned result, where the bold font highlighted words are identified medically-related entities and we concatenate them together as the processed question for retrieval.

\begin{table}[hptb]
\centering
\small
\begin{tabular}{>{\arraybackslash} m{8.1cm}}
\toprule
\textbf{Question} \\ \midrule
A \textbf{61-year-old man} presents to the \textbf{emergency department} because he has developed \textbf{blisters} at \textbf{multiple locations} on his \textbf{body}. He says that the \textbf{blisters} appeared \textbf{several days ago after} a day of hiking in the \textbf{mountains with} his colleagues. When asked \textbf{about potential} triggering \textbf{events}, he says that he recently had an \textbf{infection} and was treated \textbf{with antibiotics} but he cannot recall the \textbf{name} of the \textbf{drug} that he took. In addition, he accidentally confused his \textbf{medication with one} of his wife's \textbf{blood thinner pills several days before} the \textbf{blisters} appeared. On \textbf{examination}, the \textbf{blisters} are \textbf{flesh-colored}, raised, and \textbf{widespread} on his \textbf{skin} but do \textbf{not} involve his \textbf{mucosal surfaces}. The \textbf{blisters} are \textbf{tense} to \textbf{palpation} and do \textbf{not separate with rubbing}. \textbf{Pathology} of the \textbf{vesicles} show that they continue \textbf{under the level of the epidermis}. Which of the following is the most \textbf{likely cause} of this \textbf{patient's blistering}?
\\ \bottomrule
\end{tabular}
\caption{An example showing the extracted medically-related entities (highlighted in bold font) by MetaMap.}
\label{tab:metamap-example}
\end{table}

\section{Failure Patterns of IR}

In Table \ref{tab:error}, we showcase examples for each of the two failure patterns mentioned in the main text. 

In this table, the first example illustrates the first failure pattern, where the model fails to perform the correct one-step reasoning for disease diagnosis since the symptoms and patient history mentioned in the case description are very common symptoms (e.g. cough with sputum, increased urinary frequency) and are not disease- or condition-specific. Although the better option for this case is to examine the patient's abdomen due to his age and the lacking information of his abdominal condition, the IR system can only return some miscellaneous evidences about (1) a similar case, (2) lung cancer and smoking, (3) urinary tract infection, and (4) cholesterol issue, which can not actually answer the given question.

The second example for explaining the second failure pattern is a case about a mother with gestational diabetes / hypertension and her baby. In this case, the IR system can identify the medical condition of the mother and retrieve relevant information to this mother. However, the question is actually asking about the baby, which is only described in the very end of this question. Unfortunately, the IR system only returns the evidence for the first-step reasoning (mother's diagnosis) but not for the second-step (diagnose the baby's disease based on the mother's diagnosis).

\begin{table*}[tb]
\centering
\small
\begin{tabular}{>{\arraybackslash} m{1.1cm} |>{\arraybackslash} m{13.9cm}}
\toprule
\textbf{Question}                  & A 67-year-old man presents to a primary care clinic to establish care after moving from another state. According to his prior medical records, he last saw a physician 4 years ago and had no significant medical problems at that time. Records also show a normal EKG and normal colonoscopy results at that time. The patient reports feeling well overall, but review of systems is positive for 1 year of mild cough productive of clear sputum and 2 years of increased urinary frequency. He denies fever, chills, dyspnea, dysuria or hematuria. He denies illicit drug use but has been drinking approximately 1-2 beers per night and smoking 1 pack of cigarettes per day since age 20. Physical exam is unremarkable. Which of the following tests is indicated at this time?                                                                           \\ \midrule
\textbf{Options}                   &  \textbf{A: Abdominal ultrasound}, B: Bladder ultrasound, C: Colonoscopy, D: Serum prostate specific antigen (PSA) testing, E: Sputum culture \\ \midrule
\textbf{Evidence}                  & 
1. ... This patient is a 67-year-old man with weight loss of 10 pounds in 4 weeks and a 35 pack-year history of cigarette smoking. He quit smoking 10 year ago. He had left shoulder pain for 4 months with no dyspnea, cough, hemoptysis, or other symptoms. Massage and other musculoskeletal manipulation did not improve his symptoms. ... \\
& 2. Lung cancer, which was rare prior to 1900 with fewer than 400 cases described in the medical literature, is considered a disease of modern man. ... Tobacco consumption is the primary cause of lung cancer, ... Given the magnitude of the problem, it is incumbent that every internist has a general knowledge of lung cancer and its management. \\
& 3. The symptoms and signs of UTI vary markedly with age. ... \\
& 4. Cigarette smoking is a well-established risk factor in men and probably accounts for the increasing incidence and severity of atherosclerosis in women. ... \\
& 5. Smoking within 30 minutes of waking, smoking daily, smoking more cigarettes per day, and waking at night to smoke are associated with tobacco use disorder. ... Serious medical conditions, such as lung and other cancers, cardiac and pulmonary disease, perinatal problems, cough, shortness of breath, and accelerated skin aging, often occur. \\
& 6. Alcohol and cigarette smoking are well-known modifiers of cholesterol. ... \\ \midrule\midrule
\textbf{Question}                  & A 37-year-old G1P1001 delivers a male infant at 9 pounds 6 ounces after a C-section for preeclampsia with severe features. The mother has a history of type II diabetes with a hemoglobin A1c of 12.8\% at her first obstetric visit. Before this pregnancy, she was taking metformin, and during this pregnancy, she was started on insulin. At her routine visits, her glucose logs frequently showed fasting fingerstick glucoses above 120 mg/dL and postprandial values above 180 mg/dL. In addition, her routine third trimester culture for group B Streptococcus was positive. At 38 weeks and 4 days gestation, she was found to have a blood pressure of 176/103 mmHg and reported a severe headache during a routine obstetric visit. She denied rupture of membranes or vaginal bleeding. Her physician sent her to the obstetric triage unit, and after failure of several intravenous doses of labetalol to lower her blood pressure and relieve her headache, a C-section was performed without complication. Fetal heart rate tracing had been reassuring throughout her admission. Apgar scores at 1 and 5 minutes were 7 and 10. After one hour, the infant is found to be jittery; the infant's temperature is 96.1 F (35.6 C), blood pressure is 80/50 mmHg, pulse is 110/min, and respirations are 60/min. When the first feeding is attempted, he does not latch and begins to shake his arms and legs. After 20 seconds, the episode ends and the infant becomes lethargic. Which of the following is the most likely cause of this infant’s presentation?                                                                               \\ \midrule
\textbf{Options}                   & A: Transplacental action of maternal insulin, \textbf{B: $\beta$-cell hyperplasia}, C: Neonatal sepsis, D: Inborn error of metabolism, E: Neonatal encephalopathy \\ \midrule
\textbf{Evidence}                  & 
1. At Parkland Hospital, women with diabetes are seen in a specialized obstetrical clinic every 2 weeks. During these visits, glycemic control records are evaluated and insulin adjusted. ... \\
& 2. Once pregnancy is established, glucose control should be managed more aggressively than in the nonpregnant state. In addition to dietary changes, this enhanced management requires more frequent blood glucose monitoring and often involves additional injections of insulin or conversion to an insulin pump. ... \\
& 3. 2-hour postprandial values of 100 to 120 mg/dL, and mean he action proiles of commonly used short-and long-termdaily glucose concentrations $<$ 110 mg/dL. ... \\
& 4. The gold standard for diagnosis of insulinoma is the 72-hour monitored fast. ... \\
& 5. Treatment of gestational diabetes with a two-step strategy—dietary intervention followed by insulin injections if diet alone does not adequately control blood sugar ... \\
& 6. The rate of vertical transmission is reduced to less than 8\% by chemoprophylaxis with a regimen of zidovudine to the mother (100 mg five times/24 hours orally) started by 4 weeks gestation, ...
\\\bottomrule
\end{tabular}
\caption{\label{tab:error} Two examples of failure patterns for the IR system. The first example demonstrates the failure of capturing the correct one-step reasoning, and the second example shows the failure of focusing on the question target while correctly identifying the medical condition of the given case. The correct option is marked by the bold font. The top 6 retrieved paragraphs are shown here.}
\end{table*}

\end{document}